\documentclass[conference,onecolumn]{IEEEtran}
\IEEEoverridecommandlockouts
\usepackage{cite}
\usepackage[center]{subfigure}
\usepackage{amsmath,amssymb,amsfonts}
\usepackage{algorithmic}
\usepackage{graphicx}
\usepackage{textcomp}
\graphicspath{{img/}}
\def\BibTeX{{\rm B\kern-.05em{\sc i\kern-.025em b}\kern-.08em
    T\kern-.1667em\lower.7ex\hbox{E}\kern-.125emX}}

\usepackage[dvipsnames]{xcolor}

\usepackage{hyperref}  
\hypersetup{
  colorlinks,
  citecolor=Violet,
  linkcolor=Green,
  urlcolor=Blue
  }

\begin{document}

\title{Sensorimotor learning for artificial body perception}

\author{\IEEEauthorblockN{German Diez-Valencia, Takuya Ohashi, Pablo Lanillos*, Gordon Cheng}
\IEEEauthorblockA{\textit{Department of Electrical Engineering and Computer Sciences} \\
\textit{Technical University of Munich}\\
Munich, Germany \\
*p.lanillos@tum.de }
\thanks{This work was supported by SELFCEPTION project (www.selfception.eu) European Union Horizon 2020 Programme (MSCA-IF-2016) under grant agreement no. 741941. Workshop on Crossmodal Learning for Intelligent Robotics. IEEE Int. Conference on Intelligent Robots and Systems (IROS 2018)}
}

\maketitle






\section{Introduction}
Artificial self-perception is the machine ability to perceive its own body, i.e., the mastery of modal and intermodal contingencies of performing an action with a specific sensors/actuators body configuration \cite{lanillos2016yielding}. In other words, the spatio-temporal patterns that relate its sensors (e.g. visual, proprioceptive, tactile, etc.), its actions and its body latent variables are responsible of the distinction between its own body and the rest of the world. This paper describes some of the latest approaches for modelling artificial body self-perception: from Bayesian estimation to deep learning. Results show the potential of these free-model unsupervised or semi-supervised crossmodal/intermodal learning approaches. However, there are still challenges that should be overcome before we achieve artificial multisensory body perception.

\section{Hierarchical Bayesian models}
\begin{figure}[!hbtp]
\centering   
\includegraphics[width=0.90\columnwidth, height=80px]{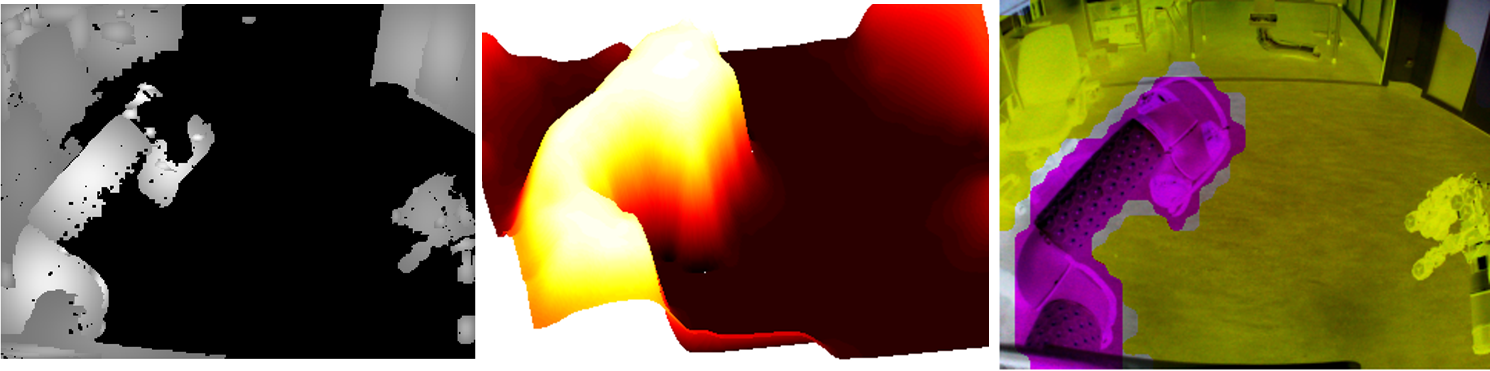}
\caption{Self-detection combining visual attention and Bayesian filtering \cite{lanillos2016yielding}. (left) Saliency segmentation; (middle) inference of the body parts; and (right) inbody vs outbody sources in the visual field.}
\label{fig:self}
\end{figure}

A first approach on self-perception was integrating multimodal tactile, proprioceptive and visual cues \cite{lanillos2016yielding} by means of Hierarchical Bayesian models and signal processing, extending \cite{gold2009using} and \cite{stoytchev2011self} ideas. Results showed that the robot was able to discern between inbody and outbody sources without using markers or simplified segmentation. Figure \ref{fig:self} shows the proto-object saliency system \cite{lanillos2015saliency3D} used as visual input and the computed probability of the image regions belonging to the robot arm. Body perception was formalized as an inference problem while the robot was interacting with the world. In order to infer which parts of the scene belong to the robot we integrated visual and accelerometers information. We defined the visual receptive field as a grid where each node (i.e., the decimation of the pixel-wise image) should be decided whether it belongs to the body or not. For that purpose, we adapted Bayesian inference grids to estimate the probability of being its body along the time. The prediction step was computed by learning the pixel-wise velocity in four directions (i.e., up, down, left, right). Furthermore, this method was successfully applied to simplify the problem of discovering objects by interaction \cite{lanillos2016self}.

\section{Predictive coding models}
\label{sec:pc}

A biologically plausible body perception model based on predictive processing \cite{friston2005theory} was also proposed in \cite{lanillos2018adaptive}. Here, body perception was transformed into approximating the latent space distribution $q(\mu)$ that defines the body schema to the real process distribution with the sensory information (posterior) $p(x|s)$. In this approach, the forward sensory model $s=g(\mu)+z$ for each modality was learnt using Gaussian process regression. Sensory fusion was computed by means of inference approximation of the body latent variables $\mu$ minimizing the free-energy bound~\cite{friston2005theory}. Furthermore, with this model, the authors were able to replicate the proprioceptive drift pattern of the rubber-hand illusion on a robot with visual, proprioceptive and tactile sensing capabilities \cite{hinz2018drifting}. 

\begin{figure*}[!t]
\centering
\subfigure[Sensory data acquisition]{\includegraphics[width=0.65\textwidth, height=90px]{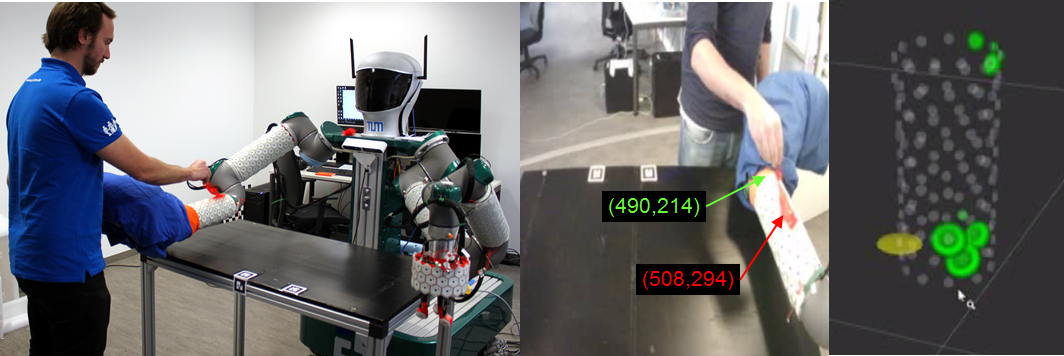} 
\label{fig:setup:a1}}
\subfigure[Rubber-hand illusion test]{\includegraphics[width=0.30\textwidth, height=90px]{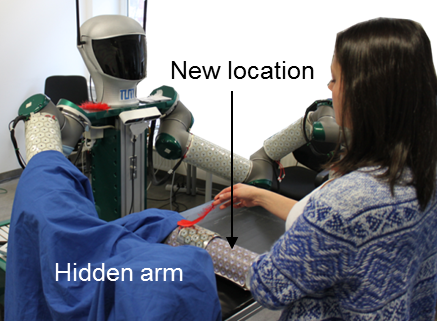}
\label{fig:setup:a2}}
\caption{Body learning and estimation through predictive coding. (a) Gathering proprioceptive (joint angles), visual robot (green) and other (red) end-effector pixel coordinates and tactile sensory data from the robot (proximity values) for different arm configurations.  Green circles represent the likelihood of being touched. (b) Adaptation test where we change the visual location of the arm and we induce synchronous visuo-tactile perturbations.}
\label{fig:setup}
\end{figure*}

\section{Generative adversarial networks (GANs)}
In order to generalize the features used in the previous predictive coding approach, we investigated deep neural networks architectures for learning the generative functions and the cross-modal relations. The advantage of using GANs as a model for self-perception is that the discriminator is a potential self/other distinction mechanism learnt in a unsupervised manner.

\subsection{Visual forward model learning}

\begin{figure}[!hbtp]
\centering   
\subfigure[Robot and network architecture]{\includegraphics[width=0.90\columnwidth,height=120px]{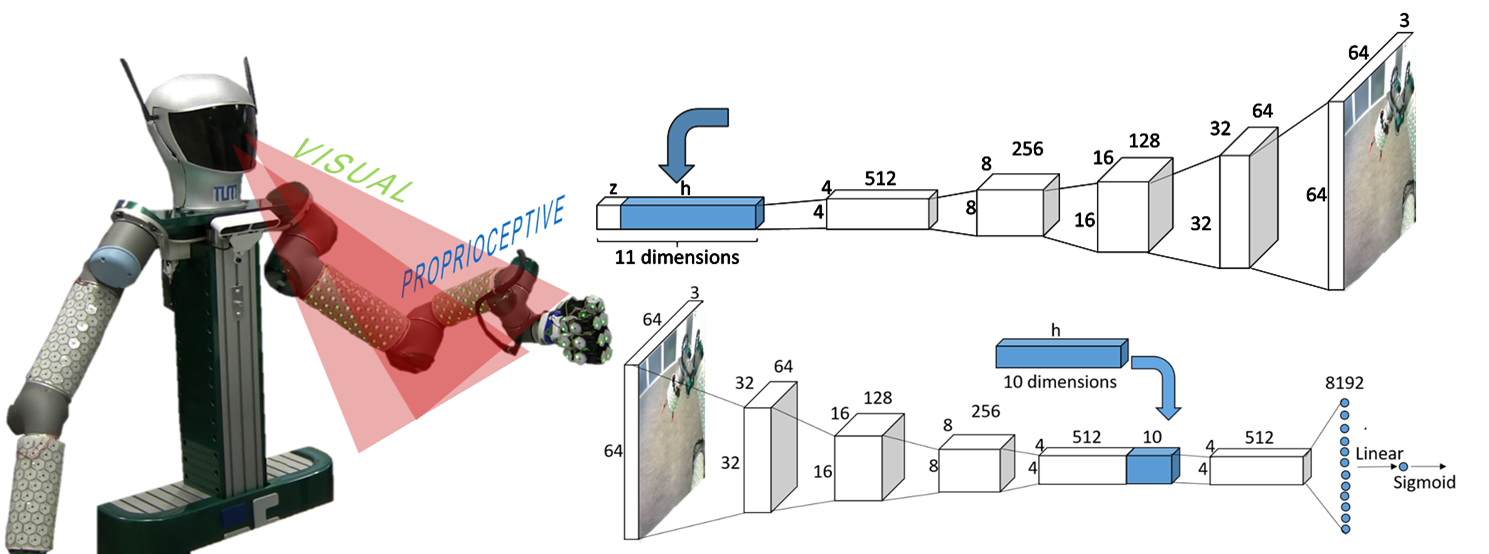}\label{fig:GAN:a}}\\
\subfigure[Training with synthetic generated data]{\includegraphics[width=0.90\columnwidth, height=75px]{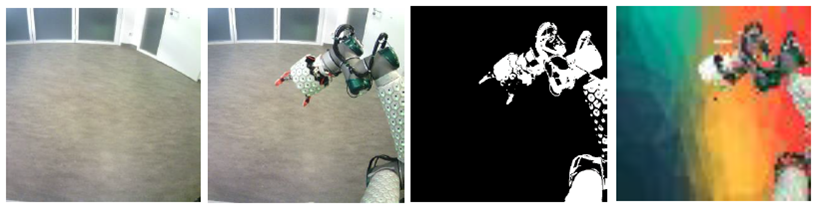} \label{fig:GAN:b}}\\
\subfigure[Generated arm visual appearance for four different joint angles configuration]{\includegraphics[width=0.90\columnwidth, height=75px]{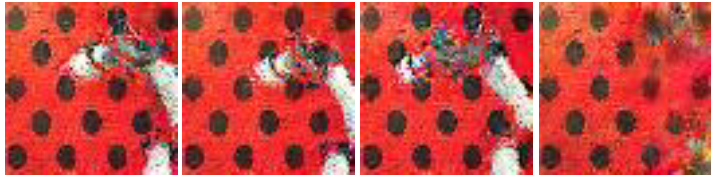} \label{fig:GAN:c}}
\caption{Forward visual-kinematic model learning using a DC-GAN. (a) Robot used and GAN architecture. (b) Example of training data generation with synthetic backgrounds. (c) Generated image from different joint arm angles configurations and unlearned background type.}
\label{fig:GAN}
\end{figure}

We analysed how the forward function that relates the joint angles (body) and the visual input can be learnt using GANs. Visuomotor learning has been also approached with recurrent neural networks in \cite{hwang2017predictive}. Conversely, here we employ a Deep Convolutional Generative Adversarial Network (DC-GAN) \cite{goodfellow2014generative,reed2016generative}. This method literally generated the arm visual shape depending on the joint angles. Figure \ref{fig:GAN:a} shows the network architecture and the robot used to extract the data. 

The great challenge was to generalize the reconstruction of the arm for any background without using segmentation. For that purpose, several background images were synthetically generated and were overlaid by automated labelled masks  (i.e., boolean mask of the arm in the visual field) by means of background subtraction (Fig. \ref{fig:GAN:b}). An example of the results of the generated arm given the a joint angle configuration is shown in Fig. \ref{fig:GAN:c}. The most right generated image shows difficulties of the model to properly reconstruct the robot arm when the majority of it is outside the field of view. Anyhow, the statistical evaluation of the network, over all experiments, showed an accuracy of 84.4\% when comparing the matching between the original versus the generated image mask.


\subsection{Cross-modal learning}
\begin{figure}[!hbtp]
\centering   
\includegraphics[width=0.90\columnwidth, height=150px]{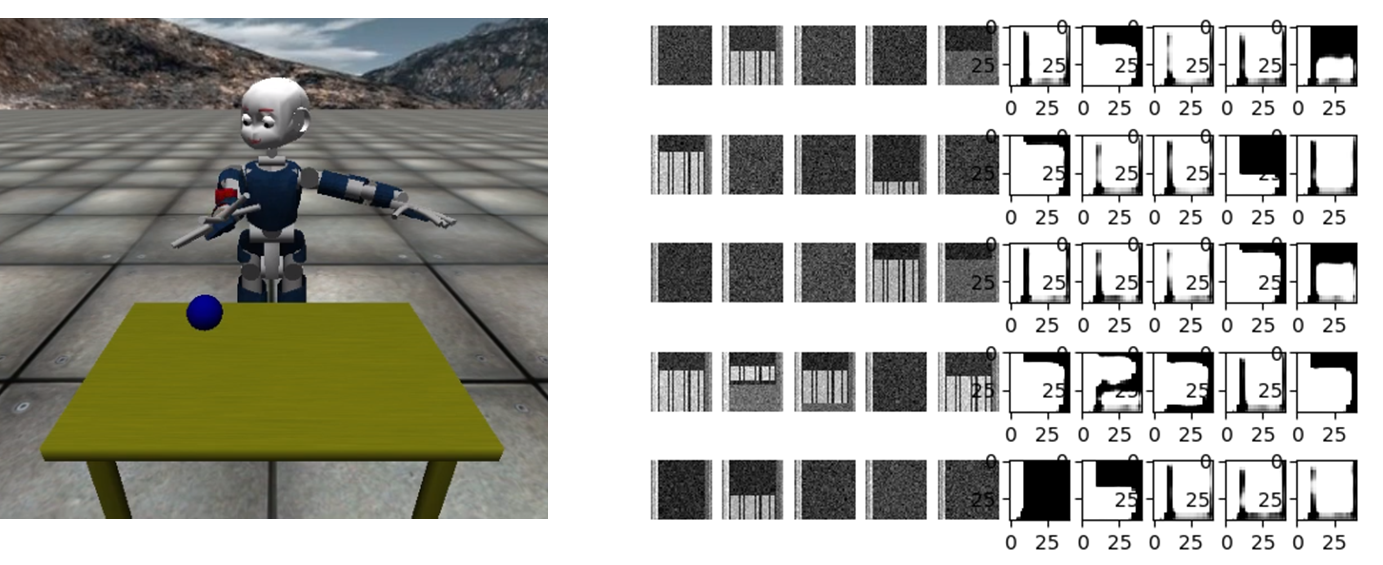}
\caption{Learning spatio-temporal multisensory patterns using denoising autoencoders. (left) simulator used to generate the data; (right) results: columns 1-5 original multimodal data, columns 6-10 reconstructed data.}
\label{fig:icubself}
\end{figure}

We further analysed self-perception from the cross-modal point of view. Instead of generating the body visual appearance from the joints angles, we extended the architecture to enable signal reconstruction from different sensor modalities using denoising autoencoders \cite{ngiam2011multimodal} but without shared representation. We used the iCub simulator \cite{tikhanoff2008open} to generate the following visuo-tactile-proprioceptive data: the left arm joint angles, the activation of the skin sensors on the left hand and left forearm and the 2D position of a red cube in the robot left eye image plane. The detection of the red cube was performed by colour blob segmentation. The skin sensors delivered a fixed length array with value 255 if there was contact or 0 otherwise. The image size was fixed to $640\times480$. We generated several left arm joints configuration, where the forearm appeared in different positions in the visual field. The red cube trajectory was then computed to touch the forearm from the hand to the elbow. After a phase of data synchronization a Wasserstein type GAN \cite{arjovsky2017wasserstein} was trained off-line. Figure \ref{fig:icubself} shows the experimental setup of the simulation and the data reconstruction results. Convolution operators were able to extract the inherent structure of the skin data but not reliable enough to provide accurate tactile reconstruction as those operators were thought for images only.

\section{Conclusion}
We have presented body learning and perception as one of the most representative and challenging cross-modal learning applications. In particular, self-perception in robots has direct applications on adaptability, safety and human-robot interaction. Through examples, we identified at least three important characteristics for modelling artificial body perception: (1) body latent space estimation through noisy sensorimotor fusion; (2) cross-modal signal recovering from multimodal information; and (3) unsupervised self-generated patterns classification. Accordingly, we have shown different techniques for partially solving the problem, such as hierarchical Bayesian models for self-detection on the visual field \cite{lanillos2016yielding}, predictive processing with GP regression for body estimation \cite{lanillos2018adaptive}, and deep nets for learning the forward visual-kinematics or visuo-tactile-proprioception relations. Further research will focus on developing a full cross-modal architecture able to properly tackle the nature of the different modalities and allowing sensor relevance tuning.





\bibliographystyle{IEEEtran}
\bibliography{pl,ninaRHI,selfception}  
\end{document}